\documentclass{article}

\usepackage{arxiv}

\usepackage[utf8]{inputenc} 
\usepackage[T1]{fontenc}    
\usepackage{hyperref}       
\usepackage{url}            
\usepackage{booktabs}       
\usepackage{amsfonts}       
\usepackage{nicefrac}       
\usepackage{microtype}      
\usepackage{lipsum}
\usepackage{graphicx}

\usepackage{times}
\usepackage{amsmath}
\usepackage{url}
\usepackage{amsthm}
\usepackage{algorithmic}
\usepackage{algorithm}
\usepackage{amssymb}
\usepackage{graphicx}
\usepackage{epsfig}
\usepackage{subfigure}
\usepackage{lipsum}
\usepackage{footnote}
\usepackage{enumitem}
\usepackage{array}
\usepackage{multirow}
\usepackage{wrapfig,lipsum,booktabs}
\usepackage{xcolor}

\newtheorem{theorem}{Theorem}

\begin{document}
\title{Learning Representations from Dendrograms}

\author{Morteza Haghir Chehreghani  \\
                Chalmers University of Technology\\
                SE-412 96 Gothenburg, Sweden\\
                \texttt{morteza.chehreghani@chalmers.se}\\
           \And
                Mostafa Haghir Chehreghani  \\
                Amirkabir University of Technology (Tehran Polytechnic)\\
                Tehran, Iran\\
                \texttt{mostafa.chehreghani@gmail.com}\\
}

\maketitle

\begin{abstract}
We propose unsupervised representation learning  and feature extraction from dendrograms.
The commonly used Minimax distance measures correspond to building a dendrogram
with single linkage criterion, with defining specific forms of a level function
and a distance function over that. Therefore, we extend this method to arbitrary dendrograms.
We develop a generalized framework wherein different distance measures and representations
can be inferred from different types of dendrograms, level functions and distance
functions. Via an appropriate embedding, we compute a vector-based representation
of the inferred distances, in order to enable many numerical machine learning algorithms
to employ such distances. Then, to address the model selection problem, we
study the aggregation of different dendrogram-based distances respectively in solution
space and in representation space in the spirit of deep representations. In the first
approach, for example for the clustering problem, we build a graph with positive and
negative edge weights according to the consistency of the clustering labels of different
objects among different solutions, in the context of ensemble methods. Then,
we use an efficient variant of correlation clustering to produce the final clusters. In
the second approach, we investigate the  combination of different distances
and features sequentially in the spirit of multi-layered architectures to obtain the final
features. Finally, we demonstrate the effectiveness of our approach via several
numerical studies.
\end{abstract}

\section{Introduction}
Real-world datasets often consist of complex and a priori unknown patterns and structures, requiring to improve the basic representation. Kernel methods are commonly used for this purpose \cite{Hofmann06areview,KernelbookShaweTaylor}. However, their applicability is confined by several limitations \cite{Luxburg:2007,Nadler07,ChehreghaniAAAI2017}. i) Finding the optimal parameter(s) of a kernel function is often nontrivial, in particular in an unsupervised learning task such as clustering where no labeled data is available for cross-validation. ii) The proper values of the parameters usually occur inside a very narrow range that makes cross-validation critical, even in presence of labeled data.

To overcome such challenges, some graph-based distance measures have been developed in the context of algorithmic graph-theory. In this setup, each object corresponds to a node in a graph, and the edge weights are the pairwise (e.g., squared Euclidean) distances between the respective objects (nodes). Then, different methods perform different types of inference on the graph to compute an effective distance measure between the pairs of objects.
Link-based methods \cite{Chebotarev:2011,YenSMS08} first sum the edge weights on every path to compute the \emph{path-specific} distances. The final distance is then obtained by summing up the \emph{path-specific} distances  of all paths between the two nodes.
This distance measure can be obtained  by inverting the Laplacian of the base distance matrix related to Markov diffusion kernel \cite{Fouss:2012,YenSMS08}. It requires an $\mathcal O(n^3)$ runtime, with $n$ the number of objects.

Minimax distance measure is an alternative option  that computes the minimum largest gap of all possible paths between the objects. Several previous works study the superior performance of Minimax distances,  compared to metric learning or link-based choices \cite{FarniaT16,FischerRB03NIPS,ChehreghaniSDM16,KimC07icml,KimC13AAAI,NIPS2011_4218,pmlrSingh}. Minimax distances have been first used with clustering problems in two ways, either as an input in the form of pairwise distance matrix \cite{ChangY08,Pavan:2007}, or integrated with some clustering algorithms \cite{FischerB03}.
The straightforward approach to compute the pairwise Minimax distances is to use an adapted variant of the Floyd-Warshall algorithm, whose runtime is $\mathcal O(n^3)$~\cite{Aho:1974}.
However, the method in \cite{FischerB03} is computationally even more demanding, as its runtime is $\mathcal O(n^2|E|+n^3\log n)$ ($|E|$ is the number of edges in the graph).
Based on equivalence of Minimax distances over a graph and over any minimum spanning tree constructed on that, \cite{ChehreghaniAAAI2017,Chehreghani2020MLj} propose to compute first a minimum spanning tree (e.g., using Prim's algorithm) and then obtain the Minimax distances over that via an efficient dynamic programming algorithm. Then, the runtime of computing pairwise Minimax distances reduces to $\mathcal O(n^2)$.  \cite{Chehreghani17ICDMminimax} analyzes computing pairwise Minimax distances in different sparse and dense settings.  \cite{ZhongMMF15} develops an approximate minimum spanning tree algorithm and investigates it for efficient computation of pairwise Minimax distances.  \cite{YuXMHXLL14,abs-1909-07774}  combine  Minimax distances with specific clustering methods in  closed-form ways.

Minimax distances have been also used for $K$-nearest neighbor search \cite{KimC07icml,KimC13AAAI,ChehreghaniSDM16}.
The method in~\cite{KimC07icml} presents a message passing method related to the sum-product algorithm~\cite{Kschischang:2006} to perform $K$-nearest neighbor classification with Minimax distances. Even though its runtime is $\mathcal O(n)$, it needs computing a minimum spanning tree (MST) in advance that can require $\mathcal O(n^2)$ runtime. Thereafter, the  algorithm in~\cite{KimC13AAAI} computes the Minimax $K$ nearest neighbors via space partitioning  whose runtime is $\mathcal O(\log n + K\log K)$. However, it is applicable only to sparse graphs built in Euclidean  spaces. Finally,~\cite{ChehreghaniSDM16} has  proposed an efficient Minimax $K$-nearest neighbor search method applicable to general graphs and dissimilarities. Its runtime, similar to the standard $K$ nearest neighbor search is  linear in general. Moreover, the method provides an outlier detection mechanism alongside performing $K$-nearest neighbor search, all with a linear runtime. The work in~\cite{ChehreghaniECIR17} investigate Minimax  $K$ nearest neighbor search for matrix (of user profiles) completion.

Besides Minimax distances, another related line of  research has been developed in the context of  \emph{tree preserving embedding} \cite{ShiehHA11,Shieh16916}, where the goal is to compute an embedding that preserves the \emph{single} linkage dendrogram in the embedding.\footnote{Tree-based structures have been studied and analyzed in several other domain such as frequent pattern mining \cite{ChehreghaniCLR11,ChehreghaniRLC07} different from the setting in this paper.}

Both Minimax distances and tree preserving embedding correspond to computing a set of features representing \emph{single} linkage dendrograms.
Therefore, this limitation motivates us to extend the previous works on representation learning and feature extraction based on \emph{single} linkage criterion and  develop a generalized framework to compute different distance measures according to various dendrograms. In our framework the dendrogram, i.e.,  the way the inter-cluster distances  called \emph{linkage} are defined, can  be constructed according to different criteria. The \emph{single} linkage criterion \cite{sneath1957dn09j} defines the linkages as the distance between the nearest members of the nodes. In contrast, the \emph{complete} linkage criterion~\cite{rensen1948method,lance67hierarchical} defines the distance between two nodes as the distance between their farthest members, which corresponds to the maximum within-node distance of the new node. On the other hand, in \emph{average} criterion~\cite{sokal58} the average of inter-node distances is used as the linkage between two nodes. The \emph{Ward} method ~\cite{Inchoate:Ward63} uses the distances between the means of the nodes normalized by a function of the size of the nodes. \cite{MoseleyW17} analyzes in detail several of such criteria.

We  study the embedding of the pairwise distances computed from a dendrogram into a new vector space such that the squared Euclidean distances in the new space equal to the dendrogram-based distances. This embedding provides us to employ dendrogram-based distances with a wide range of different machine learning methods, and yields a rich family of alternative dendrogram-based distances with Minimax distance measures and tree preserving embedding in \cite{ShiehHA11,Shieh16916} being only  special instantiations.

Then, we encounter a model selection problem which asks for the choice of the appropriate distance measure (and dendrogram). Therefore,  in the context of model averaging and ensemble methods, we first study the aggregation of the distance measures from different dendrograms in the solution space. Assuming, for exaple the different dendrogram-based distance measures are used for an unsupervised clustering task, we build a graph with positive and negative edge weights based on the (dis)agreement of the respective nodes among different clustering solutions. Then, we employ an efficient variant of correlation clustering to obtain the final ensemble solution.
Second, several recent studies demonstrate the superior performance of deep representation learning models that extract complex features via aggregating representations sequentially at different levels. Such models are highly over-parameterized and thus require huge amounts of training data to infer the parameters. However, unsupervised representation learning is expected to become far more important in longer term, as human and animal learning is mainly unsupervised~\cite{LeCunBH15Nature}. Thereby, with the possibility of having access to a wide range of alternative feature extraction models, we investigate  design of multi layer deep architectures in an unsupervised manner (in representation space, instead of solution space) which does not require inferring or fixing any critical parameter. Specifically, we study the sequential aggregation of the dendrogram-based features where for example the \emph{single} linkage features are computed based on the features obtained from \emph{average} linkage dendrogram, instead of using the original data features.

Our framework provides several options for choosing the dendrogram and the level function, where each option yields separate unsupervised representations and features. However, at the same time, we propose  a principled way to aggregate and choose the best options (either in solution space or in representation space).
Availability of such alternatives endows a rich family of unsupervised representation learning methods and is different from optimizing the free parameters of a kernel. We will discuss this model selection aspect with more detail in the experiments section.

Finally, we experimentally validate  the effectiveness of our framework on UCI and real-world datasets.

\section{Feature Extraction from Dendrograms}

In this section, we first introduce the setup for computing distance measures from dendrograms, and then, based on the relation between Minimax distances and \emph{single} linkage agglomerative clustering, we propose a generalized approach to extract features from dendrograms.

\subsection{Pairwise distances over dendrograms}
We are given a dataset of $n$ objects with indices $\mathbf O= \{1,...,n\}$ and the corresponding measurements. The measurements can be for example the vectors in a feature space  or the pairwise distances between the objects. In the former case, the measurements are shown by the $n\times d$ matrix $\mathbf Y$, wherein the $i^{th}$ row (i.e., $\mathbf Y_i$) specifies the $d$ dimensional vector of the  $i^{th}$ object. In the latter form, an $n \times n$ matrix $\mathbf X$ represents the pairwise distances between the objects. Then, we might show the data by graph $\mathcal G(\mathbf O,\mathbf X)$, wherein $\mathbf O$ is the set of its vertices and $\mathbf X$ represents the edge weights.
Note that the former is a specific form of the latter representation, where the pairwise distances are computed according to (squared) Euclidean distances.

A dendrogram $D$ is defined as a rooted ordered tree such that,
\begin{enumerate}
\item each node $v$ in $D$ includes a non-empty subset of the objects, i.e., $v \subset \mathbf O,|v| >0, \forall v \in D$, and
\item  the overlapping nodes are ordered, i.e., $\forall u,v \in D, \text{if } u \cap v \ne 0, \text{ then either } u \subseteq v \text{ or } v \subseteq u$.
 \end{enumerate}

 The latter condition implies that between every two overlapping nodes an ancestor-descendant relation holds, i.e., $u \subseteq v$ indicates $v$ is an ancestor of $u$, and $u$ is a descendant of $v$.

The nodes at the lowest level (called the \emph{final} nodes) are the singleton objects, i.e., node $v$ is a final node if and only if $|v|=1$.  A node at a higher level contains the union of the objects of its children (direct descendants). The root of a dendrogram is defined as the node at the highest level (which has the maximum size), i.e., all other nodes are its descendants. $linkage(v)$ returns the distance between the children of $v$ based on the criterion used to compute the dendrogram. For the simplicity of  explanation, we assume each node has only two children. In the case that a parent node has multiple (more than two) child nodes, the different linkages among the children will have the same value, which will be assigned to the parent node. To encode a dendrogram, we use the data structure supported by SciPy in Python in particular the same way as the output of the \emph{linkage} function.\footnote{scipy.cluster.hierarchy.linkage: \url{https://docs.scipy.org/doc/scipy/reference/generated/scipy.cluster.hierarchy.linkage.html}}
This data structure is a $n-1$  by $4$ matrix called $\mathbf Z$. Each individual object constitutes a separate singleton cluster where the cluster index is the object index.  At each iteration $i$ of the agglomerative algorithm, the indices of the two combined clusters are stored respectively in $\mathbf Z_{i,0}$ and $\mathbf Z_{i,1}$. The index of the new cluster is then $i+n$. We store the distance between the two clusters in $\mathbf Z_{i,2}$ and the size of the new cluster in $\mathbf Z_{i,3}$.

The level of node $v$, i.e., $level(v)$ is determined by  $\max(level(c_l), level(c_r))+1$, where $c_l$ and $c_r$ indicate the two child nodes of $v$. For the final nodes, the $level()$ function returns $0$.
Every connected subtree of $D$ whose final nodes contain only singleton objects from $\mathbf O$ constitutes a dendrogram on this set. We use $\mathcal D^D$ to refer to the set of all (sub)dendrograms derived in this form from $D$.

Thereby, the level of node $v$, i.e., $level(v)$ is determined by
\begin{equation}
    level(v) =
\begin{cases}
    \max(level(c_l), level(c_r))+1,& \\ \quad\quad\quad\text{if } linkage(v) > \max(linkage(c_l), linkage(c_r)).\\
    \max(level(c_l), level(c_r)),    &\\ \quad\quad\quad\text{if } linkage(v) = \max(linkage(c_l), linkage(c_r)).
\end{cases}
\end{equation}

Where $c_l$ and $c_r$ indicate the two child nodes of $v$.
Note that in an agglomerative method we always have $linkage(v) \ge \max(linkage(c_l,c_r))$. In particular,  we usually expect $linkage(v) > \max(linkage(c_l,c_r))$, unless there are ties for example in the case of \emph{single} linkage method, where then the new combination does not yield a higher level node. Rather, the new node has effectively three children instead of two, where two of them are combined to make an intermediate node. Without loss of generality and for the sake of simplicity of presentation, we assume that ties do not occur, i.e., we always have
\begin{equation}
    level(v) = \max(level(c_l,c_r))+1.
\label{eq:basic_level}
\end{equation}

We consider a generalized variant of the $level()$ function over a dendrogram $D$. Any function $f(v)$ that satisfies the following conditions is a \emph{generalized level} function.

\begin{enumerate}
\item $f(v) = 0$ if and only if $v \subset \mathbf O , |v| =1$.\\
\item $f(v) > f(u)$ if and only if $v$ is an ancestor of $u$.
\end{enumerate}

It is obvious that the basic function $level()$ satisfies these conditions.  We use $v^*_{ij}$ to denote the node at the lowest level which contains both $i$ and $j$, i.e.,

\begin{equation}
	v^*_{ij} = \arg \min_{v\in D} f(v) \quad \text{ s.t. } i,j \in v.
\end{equation}

Given dendrogram $D$, each node $v \in D$ represents the root of a dendrogram $D' \in \mathcal D^D$. Thereby, the dendrogram $D'$ inherits the properties of its root node, i.e., $f(D') = \max_{v\in D'} f(v)$ and $linkage(D') = \max_{v\in D'} linkage(v)$, since the root node has the maximum linkage and level among the nodes of $D'$.

In this paper, we investigate inferring pairwise distances from a dendrogram computed according to an arbitrary criterion, i.e., beyond \emph{single} linkage criterion. Moreover, our framework allows one to define the level function in a very flexible and diverse way.
For this purpose, we consider the following generic distance measure over dendrogram $D$, where $\mathbf D^D_{ij}$ indicates the pairwise dendrogram-based distance between the pair of objects (final nodes) $i, j\in \mathbf O$.
\begin{equation}
	\mathbf D^D_{ij} = \min f(D') \quad \text{s.t.} \quad i,j \in D',  \text{ and } D'\in \mathcal{D}^D.
\label{eq:generic_hierarchy_dist}
\end{equation}

The level function $f(v)$ and the distance matrix $\mathbf D^D$ provide distinguishing outliers at different levels. The outlier objects do not occur in the nearest neighborhood of many other clusters or objects. Thus, they join the other nodes of the dendrogram only at higher levels. Hence, the probability of object $i$ being an outlier is proportional to the level at which it joins to other objects/clusters. Therefore, such objects will have a large dendrogram-based distance from the other objects.

\subsection{Minimax distances and single linkage agglomeration}
We first study the relation between Minimax distances and \emph{single} linkage agglomerative method. In particular, we elaborate that given the pairwise dissimilarity matrix $\mathbf X$, the pairwise Minimax distance between objects $i$ and $j$ is equivalent to $\mathbf D^D_{ij}$ where the dendrogram is produced with \emph{single} linkage criterion and $\mathbf D^D_{ij}$ is defined by
\begin{equation}
	\mathbf D^D_{ij} = \min linkage(D') \quad \text{s.t.} \quad i,j\in D'  \text{ and } D'\in \mathcal{D}^D \, ,
\label{eq:singleMinimax}
\end{equation}
i.e., $f(D')$ in Eq.~\ref{eq:generic_hierarchy_dist} is replaced by $linkage(D')$.

\begin{theorem}
For each pair of objects $i,j \in \mathbf O$, their Minimax distance measure over graph $\mathcal G(\mathbf O,\mathbf X)$ is equivalent to their pairwise distance $\mathbf D^D_{ij}$ defined in Eq.~\ref{eq:singleMinimax} where the dendrogram $D$ is obtained according to single linkage agglomerative method.
\label{theorem:MinimaxEmbedding}
\end{theorem}

\begin{proof}
It can be shown that the pairwise Minimax distances over an arbitrary graph are equivalent to pairwise Minimax distances over `any' minimum spanning tree computed from the graph. The proof is similar to the \emph{maximum capacity} problem~\cite{Hu61} problem. Thereby, the Minimax distances are obtained by

\begin{eqnarray}
	\mathbf D_{i,j}^{MM} &=& \min_{r\in \mathcal R_{ij}(\mathcal G)}\left\{ \max_{1\le l \le |r|-1}\mathbf X_{r(l)r(l+1)}\right\} \nonumber \\
	&=&  \max_{1\le l \le |r_{ij}|-1}\mathbf X_{r(l)r(l+1)},
\label{Eq.pathTree}
\end{eqnarray}

where $r_{ij}$ indicates the (only) route between $i$ and $j$, i.e., to obtain Minimax distances $\mathbf D^{MM}_{ij}$, we select the maximal edge weight on the only route between $i$ and $j$ over the minimum spanning tree.

On the other hand, single linkage method and the Kruskal's minimum spanning tree algorithm are equivalent \cite{Gower69}.
Thus, dendrogram $D$ represents the pairwise Minimax distances. Now, we only need to show that the Minimax distances in Eq. \ref{Eq.pathTree} equal the distances defined in Eq. 3 of the main text, i.e.,  $\mathbf D^D_{ij}$ is the largest edge weight on the route between $i$ and $j$ in the hierarchy.

Given $i,j$, let $D^* = \arg\min linkage(D') \quad \text{s.t.} \quad i,j\in D'  \text{ and } D'\in \mathcal{D}^D$. Then, $D^*$ represents a minimum spanning subtree, which includes a route between $i,j$ (because the root node of $D^*$ contains both $i,j$) and it is consistent with a complete minimum spanning on all the objects. On the other hand, we know that for each pair of nodes $u, v\ \in D^*$ which have direct or indirect parent-child relation, we have, $linkage(u) \ge linkage(v)$ iff $f(u) \ge f(v)$. This indicates that the linkage of the node root of $D^*$ represents the maximal edge weight on the route between $i$ and $j$ induced by the dendrogram $D$. Thus,  $\mathbf D^D_{ij}$ defined in Eq. 3 of the main text represents $\mathbf D^{MM}_{ij}$ and the proof is complete.
\end{proof}

Notice that the Minimax distances in Eq.~\ref{eq:singleMinimax} are obtained by replacing $f(D')$ with $linkage(D')$ in the generic form of Eq.~\ref{eq:generic_hierarchy_dist}.

\subsection{Vector-based representation of dendrogram-based distances}

The generic distance measure defined in Eq.~\ref{eq:generic_hierarchy_dist} yields an $n \times n$ matrix of pairwise dendrogram-based distances between objects. However, a lot of machine learning algorithms perform on a vector-based representation of the objects, instead of the pairwise distances. For instance, mixture density estimation methods such as Gaussian Mixture Models (GMMs) fall in this category. Vectors constitute the most basic form of data representation, since they provide a bijective map between the objects and the measurements, such that a wide range of numerical machine learning methods can be employed with them. Moreover,  feature selection is more straightforward with this representation.
Thereby, we compute an embedding of the objects into a new space, such that their pairwise squared Euclidean distances in the new space equal to their pairwise distances obtained from the dendrogram.
For this purpose, we first investigate the feasibility of this kind of embedding. Theorem~\ref{theorem:HierarchyEmbedding} verifies the existence of an $\mathcal{L}_2^2$ embedding for the general distance measure defined in Eq.~\ref{eq:generic_hierarchy_dist}.\footnote{Note that $\mathbf X$  is not required to induce a \emph{metric}, i.e., the  triangle inequality might fail.}
\begin{theorem}
Given the dendrogram $D$ computed on the input data $\mathbf Y$ or $\mathbf X$, the matrix of pairwise distances $\mathbf D^D$ obtained via Eq.~\ref{eq:generic_hierarchy_dist} induces an $\mathcal{L}_2^2$ embedding, such that there exists a new vector space for the set of objects $\mathbf O$ wherein the pairwise squared Euclidean distances equal to $\mathbf D^D_{ij}$'s in the original data space.
\label{theorem:HierarchyEmbedding}
\end{theorem}
\begin{proof}
First, we show that the matrix $\mathbf D^{D}$ yields an \emph{ultrametric} \cite{Leclerc1981}. The conditions to be satisfied are:
\begin{enumerate}[leftmargin=*]
\item $\forall i,j: \mathbf D^{D}_{ij} = 0$ if and only if $i=j$. We investigate each of the conditions separately. i) First, if $i=j$, then $\mathbf D^{D}_{ii} = \min f(i) = 0$. ii) If $\mathbf D^{D}_{ij} = 0$, then $v^*_{ij} = i = j$, because $f(v) = 0$ if and only if $v \in \mathbf O$. On the other hand, $\forall i\ne j, \mathbf X_{ij} >0$, i.e., $f(v^*_{ij}) > 0$ if $i\ne j$.
\item $\forall i,j: \mathbf D^{D}_{ij} \ge 0$. We have, $\forall v, f(v) \ge 0$. Thus, $\forall D' \in \mathcal D^D, \min f(D) \ge 0$, i.e., $\mathbf D^{D}_{ij} \ge 0$.
\item $\forall i,j: \mathbf D^{D}_{ij} = \mathbf D^{D}_{ji}$. We have, $\mathbf D^{D}_{ij} = \{\min f(D) \quad \text{s.t.} \quad i,j \in D',  \text{ and } D'\in \mathcal{D}^D\} = \{\min f(D) \quad \text{s.t.} \quad j,i \in D',  \text{ and } D'\in \mathcal{D}^D\} = \mathbf D^{D}_{ji}$.
\item $\forall i,j,k: \mathbf D^D_{ij}  \le  \max(\mathbf D^D_{ik},\mathbf D^D_{kj})$. We first investigate $\mathbf D^D_{ik}$ where we consider the two following cases: i) If $\mathbf D^D_{ij} \le \mathbf D^D_{ik}$ (Figure~\ref{fig:ultra1}), then $\mathbf D^D_{ik}$ does not yield a contradiction. ii) If  $\mathbf D^D_{ij} > \mathbf D^D_{ik}$, then $i$ and $k$ join earlier than $i$ and $j$, i.e., $f(v^*_{ij}) > f(v^*_{ik})$ (Figure~\ref{fig:ultra2}). In this case, we have $f(v^*_{ij}) = f(v^*_{v^*_{ik},j})$ and $f(v^*_{kj}) = f(v^*_{v^*_{ik},j})$. Thus, we will have $f(v^*_{ij}) = f(v^*_{kj})$, i.e., $\mathbf D^D_{ij} = \mathbf D^D_{ik} \le \max(\mathbf D^D_{ik},\mathbf D^D_{kj})$. In a similar way, by investigating $\mathbf D^D_{jk}$ a similar result holds. Thereby, we conclude, a) if $\mathbf D^D_{ij} > \mathbf D^D_{ik}$, then $\mathbf D^D_{ij} = \mathbf D^D_{kj}$, and b) if $\mathbf D^D_{ij} > \mathbf D^D_{kj}$, then $\mathbf D^D_{ij} = \mathbf D^D_{ik}$. Thereby, we always have $\mathbf D^D_{ij}  \le  \max(\mathbf D^D_{ik},\mathbf D^D_{kj})$.
\end{enumerate}

On the other hand, one can show that an \emph{ultrametric} induces an $\mathcal{L}_2^2$ embedding~\cite{opac-b1048711}. Therefore, $\mathbf D^{D}$ represents the pairwise squared Euclidean distances in a  new vector space.

\begin{figure}[ht!]
    \centering
    \subfigure[$\mathbf D^D_{ij} \le \mathbf D^D_{ik}$]
    {
        \includegraphics[width=0.2\textwidth]{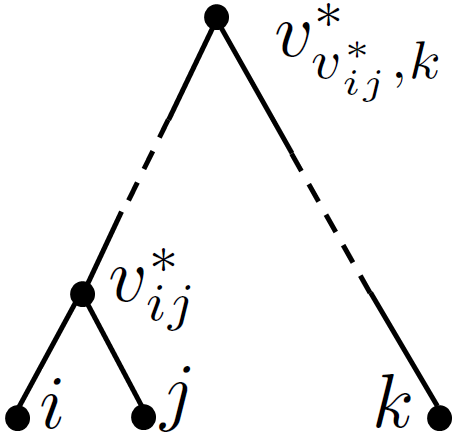}
        \label{fig:ultra1}
    }
    \hspace{13mm}
    \subfigure[$\mathbf D^D_{ij} > \mathbf D^D_{ik}$]
    {
        \includegraphics[width=0.2\textwidth]{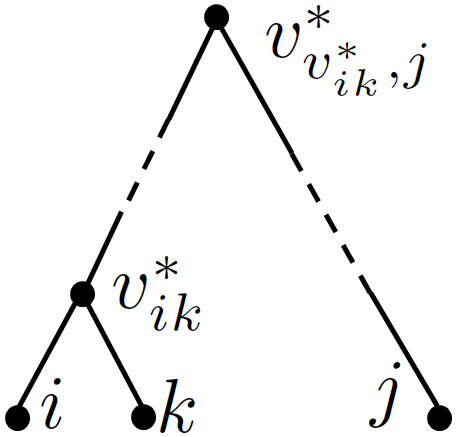}
        \label{fig:ultra2}
    }
    \caption{The \emph{ultrametric} property of $\mathbf D^D$.}
    \label{fig:Hierarchy_Ultra}
\end{figure}
\end{proof}

After assuring the existence of such an embedding, we can use any method to compute it. In particular, we exploit the method introduced in~\cite{RePEc1938} and then further analyzed in~\cite{torgerson1958theory}. This method proposes first centering $\mathbf D^{D}$ to obtain a Mercer kernel and then performing an eigenvalue decomposition:
\footnote{In~\cite{roth03PAMI}, this method has been used to obtain an $K$-means variant for pairwise clustering, after adding a large enough constant to the off-diagonal elements of the input distance matrix.}

\begin{enumerate}
\item Center $\mathbf D^{D}$ via
    \begin{equation}
   	 \mathbf{W}^{D}\leftarrow -\frac{1}{2}\mathbf{A} \mathbf D^{D} \mathbf{A}.
	\label{Eq:centering}
    \end{equation}
    $\mathbf{A}$ is obtained by $\mathbf{A} = \mathbf{I}_n - \frac{1}{n}\mathbf{e}_n\mathbf{e}_n^{T}$, where $\mathbf{e}_n$ is an $n-$dimensional constant vector of $1$'s and $\mathbf{I}_n$ is an identity matrix of size $n\times n$.

\item With this transformation, $\mathbf W^{D}$ becomes a positive semidefinite matrix. Thus, we decompose $\mathbf W^{D}$ into its eigenbasis, i.e., $\mathbf W^{D}=\mathbf{V}\boldsymbol{\Lambda}\mathbf{V}^{T},$ where $\mathbf{V} = (v_1,...,v_n)$ contains the eigenvectors $v_i$ and $\boldsymbol{\Lambda}=\texttt{diag}(\lambda_1,...,\lambda_n)$ is a diagonal matrix of eigenvalues $\lambda_1\geq...\geq\lambda_l\geq\lambda_{l+1}= 0 = ... = \lambda_n$.  Note that the eigenvalues are nonnegative, since $\mathbf W^{D}$ is positive semidefinite.

\item Calculate the $n\times l$ matrix $\mathbf{Y}^{D}_l=\mathbf{V}_l(\boldsymbol{\Lambda}_l)^{1/2},$ with $\mathbf{V}_l=(v_1,...,v_l)$ and   $\boldsymbol{\Lambda}_l=\text{diag}(\lambda_1,...,\lambda_l)$, where $l$ shows the dimensionality of the new vectors.
\end{enumerate}

The new dendrogram-based dimensions are ordered according to the respective eigenvalues and one might choose only the first most representative ones, instead of taking all. Hence, an advantage of computing such an embedding is feature selection.

\subsection{On the choice of level function}
\begin{figure}[t!]
    \centering
    \includegraphics[width=0.4\textwidth]{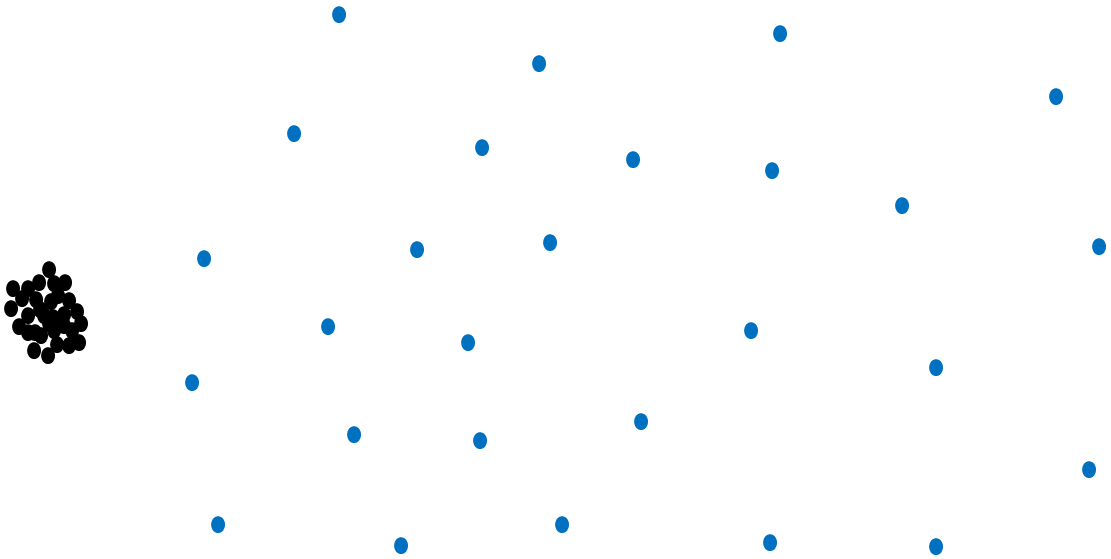}
    \caption{Minimax distance measures might perform imperfectly on the data with diverse densities.
    An adaptive approach which takes into account the variance of different classes or clusters might be more appropriate.
    }
    \label{fig:hierarchy_distance}
\end{figure}

As mentioned before, Minimax distances as a particular instance of the dendrogram-based representations, are widely used in clustering and classification tasks. However,
such distances (and equivalently the \emph{single} linkage method) do not take into account the diverse densities of the structures or classes. For example, consider the dataset shown in Figure~\ref{fig:hierarchy_distance} which consists of two clusters with different densities, marked respectively with black and blue colors. However, the intra-cluster Minimax distances for the members of the blue cluster are considerably large compared to the intra-cluster Minimax distances of the black cluster, or even the inter-cluster Minimax distances. Thereby, a clustering algorithm might split the blue cluster, instead of performing a cut on the boundary of the two clusters.
According to Proposition~\ref{theorem:MinimaxEmbedding}, the Minimax distance between objects $i$ and $j$ seeks for a linkage with maximal  weight on the path between them in the dendrogram. However, the absolute value of a linkage might be biased in a way that it does not precisely reflect the real coherence of the two nodes compared to the other nodes/objects. Thereby, in order to be more adaptive with respect to the diverse densities of the underlying structures, we will investigate the following choice in our experiments.
\begin{equation}
	\mathbf D^D_{ij} = \min_{D'} level(D') \quad \text{s.t.} \quad i,j \in D',  \text{ and } D'\in \mathcal{D}^D.
\label{eq:level_hierarchy_dist}
\end{equation}
Note that our analysis is generic and can be applied to any definition of dendrogram-based distance measure and to any choice of $f$ defined in Eq.~\ref{eq:generic_hierarchy_dist}. It only needs to satisfy the aforementioned conditions for generalized level functions.

\section{Aggregation of Multiple Representations}

\subsection{Aggregation in solution space}

As discussed earlier, a dendrogram can be constructed in several ways according to different criteria. Moreover, the choice of a level function and a distance function over a dendrogram renders another degree of freedom.
Therefore, choosing the right method constitutes a model selection question. Let us assume such distances and features are used later in a clustering task, which is the most common unsupervised learning problem. Then, we address this problem via an ensemble method in the context of model averaging.

We follow a two-step procedure to compute an aggregated clustering  that represents a given set of  clustering solutions (where, e.g., each solution is the result of a particular dendrogram and then a clustering algorithm). First, we construct a graph whose vertices represent the objects and its edge weights can be any integer number (i.e., positive, negative or zero), depending how often the respective vertices appear at the same cluster among the $M$ different clustering solutions.
More specifically, we initialize the edge weights by zero. Then, for each clustering solution $\mathbf c^m \in \{1,...,K\}^n, 1\le m\le M$ (each obtained from a different dendrogram-based representation), we compute a co-clustering matrix whose $(i,j)^{th}$ entry is $+1$ if $\mathbf c^m_i = \mathbf c^m_j$, and it is $-1$ otherwise ($K$ indicates the number of clusters). Finally, we sum up the co-clustering matrices to obtain $\mathbf S^e$. Algorithm~\ref{alg:ensemble} describes the procedure in detail.\footnote{The work in \cite{Chehreghani17ICDM} suggests an adaptive shift approach to build the correlation matrix from a given dissimilarity matrix. However, in this work the correlation matrix is given by construction.}

\begin{algorithm}[!ht]
\caption{Aggregation of $M$ clustering solution by correlation clustering.}
\label{alg:ensemble}
\begin{algorithmic} [1]
\REQUIRE {A set of $M$ clustering solutions $\mathbf c^m, 1\le m\le M$ on the same set of objects $\mathbf O$.}
\ENSURE An ensemble clustering solution $\mathbf c^e$.

\FOR {$i \in \mathbf O$}
	\FOR {$j \in \mathbf O$}
		\STATE $\mathbf S_{ij}^e = 0$
	\ENDFOR
\ENDFOR

\FOR {$1\le m\le M$}
	\FOR {$i \in \mathbf O$}
		\FOR {$j \in \mathbf O$}
			\IF {$\mathbf c^m_i = \mathbf c^m_j$}
				\STATE $\mathbf S_{ij}^e =  \mathbf S_{ij}^e +1$
			\ELSE
				\STATE $\mathbf S_{ij}^e =  \mathbf S_{ij}^e -1$
			\ENDIF
		\ENDFOR
	\ENDFOR
\ENDFOR

\STATE Apply \emph{Correlation Clustering} on $\mathbf S^e$ to obtain final clustering solution $\mathbf c^e$.

\STATE \textbf{return} $\mathbf c^e$.
\end{algorithmic}
\end{algorithm}

Given the graph with positive and negative edge weights, we use \emph{correlation clustering}~\cite{BansalBC04} to partition it into $K$ clusters. This model computes a partitioning that minimizes the disagreements, i.e., sum of the inter-cluster positive edge weights plus sum of the intra-cluster negative edge weights should be minimal. The cost function for a fixed number of clusters $K$ is written by \cite{BansalBC04,ChehreghaniBB12}
\begin{align}
    R(\mathbf c,\mathbf S^e) =& \frac{1}{2}\sum_{k=1}^{K} \sum_{i,j\in \mathbf O_k}(|\mathbf S^e_{ij}|-\mathbf S^e_{ij}) \nonumber \\
 & + \frac{1}{2}\sum_{k=1}^{K} \sum_{k'=k+1}^{K} \sum_{i\in \mathbf O_k} \sum_{j\in \mathbf O_{k'}} (|\mathbf S^e_{ij}|+\mathbf S^e_{ij}),
    \label{eq:CC}
\end{align}
where $\mathbf O_k$ indicates the objects of the $k^{th}$ cluster, i.e., $\forall i: i \in \mathbf O_k \text{ iff } \mathbf c_i =k$.  This model has been further analyzed in \cite{ThielCD19} in terms of convergence rate.

This ensemble clustering method yields a consistent aggregation of the clustering solutions obtained from different representations, i.e., in the case of $M=1$ the optimal solution of Eq.~\ref{eq:CC}  does not change the given clustering solution of this single representation.

\paragraph{\textbf{Efficient optimization of correlation clustering cost function.}}
Finding the optimal solution of the cost function in Eq.~\ref{eq:CC} is NP-hard~\cite{BansalBC04,DemaineEFI06} and even APX-hard~\cite{DemaineEFI06}. Therefore, we develop a \emph{local search} method which  computes a local minimum of the cost function. The good performance of such a greedy strategy is well studied for different clustering models, e.g., $K$-means~\cite{Macqueen67somemethods},  kernel $K$-means~\cite{Scholkopf:1998} and in particular several graph partitioning methods~\cite{Dhillon:2004,Dhillon:EtAl:05}.\footnote{Consistently, for correlation clustering we observe a better performance with the local search method compared to the different approximation schemes such as those proposed in~\cite{BansalBC04,DemaineEFI06}.} We begin with a random clustering solution and then we iteratively assign each object to the cluster that yields a maximal reduction in the cost function. We repeat this procedure until no further improvement is achieved, i.e., a local optimal solution is found.

At each step of the aforementioned procedure, one needs to investigate the costs of assigning every object to each of the clusters. The cost function is quadratic, thus, a single evaluation might take $\mathcal O(n^2)$. Thereby, if the local search converges after $t$ steps, the total runtime will be $\mathcal O(tn^3)$. However, we do not need to recalculate the cost function for each individual evaluation.
Let $R(\mathbf c,\mathbf S^e)$ denote the cost of clustering solution $\mathbf c$, wherein the cluster label of object $i$ is $k$. To obtain a more efficient cost function evaluation, we first consider the contribution of object $i$ in $R(\mathbf c,\mathbf S^e)$, i.e., $R_i(\mathbf c,\mathbf S^e)$,  which is written by
\begin{equation}
R_i(\mathbf c,\mathbf S^e) = \frac{1}{2}\sum_{j\in \mathbf O_k} (|\mathbf S^e_{ij}|-\mathbf S^e_{ij})
+ \frac{1}{2}\sum_{q=1, q\ne k}^{K}\sum_{j\in \mathbf Oq} (|\mathbf S^e_{ij}|+\mathbf S^e_{ij}) .
\label{eq:CostContibution}
\end{equation}
Then, the cost of the  clustering  solution $\mathbf c'$ being identical to $\mathbf c$ except for the object $i$ which is assigned to cluster $k' \ne k$, i.e., $R(\mathbf c',\mathbf S^e)$ is computed by
\begin{equation}
R(\mathbf c',\mathbf S^e) = R(\mathbf c,\mathbf S^e) - R_i(\mathbf c,\mathbf S^e) + R_i(\mathbf c',\mathbf S^e),
\label{eq:updateCC}
\end{equation}
where $R(\mathbf c,\mathbf S^e)$ is already known and $R_i(\mathbf c,\mathbf S^e)$ and $R_i(\mathbf c',\mathbf S^e)$ both require an $\mathcal O(n)$ runtime.
Thus, we evaluate the cost function~\ref{eq:CC} only once for the initial random clustering. Then, iteratively and until the convergence, we compute the costs of assigning objects to different clusters via Eq.~\ref{eq:updateCC} and assign them to the clusters that yields a minimal cost. The total runtime  is then $\mathcal O(tn^2)$.

\subsection{Aggregation in representation space}
In this section, instead of an ensemble-based approach in the solution space, we describe the aggregation of different (dendrogram-based) distances in the representation space, independent of what the next task will be.
The embedding phase of our general-purpose framework not only enables us to employ any numerical machine learning algorithm, but also provides an amenable way to successively combine different representations. In this approach, the features extracted from a dendrogram (e.g., single linkage) are used to build another dendrogram according to the same or a different criterion (e.g., average linkage), in order to yield more complex features. The degree of freedom (richness of the function class) can increase by the choice of a different level or distance function over dendrograms.  Such a framework leads to a \emph{nonparametric deep architecture} wherein a cascade of multiple layers of nonparametric information processing units are deployed for feature transformation and extraction. The output of each layer is a set of features, which can be fed into another layer as input. Note that in this  architecture any other (nonparametric) unit can be employed at the layers, beyond the dendrogram-based feature extraction units. Each layer (dendrogram) extracts a particular type of features in the space of data representation.

\section{Experiments}
\label{sec:experiments}

We empirically investigate the performance of dendrogram-based representations on different datasets and demonstrate the usefulness of this approach to extract suitable features. Our methods are unsupervised and do not assume availability of any labeled data. Thus, to fully benefit from this property,   we consider an unsupervised representation learning strategy, such that no free parameter is involved in inferring the new features. Thereby, we apply our methods to clustering and density estimation problems, for which parametric feature extraction methods might be inappropriate, due to lack of labeled data for cross validation (to estimate the parameters). In particular, after extracting the new features, we apply the following algorithms to obtain a clustering solution: i) Gaussian Mixture Model (GMM), ii) $K$-means, and iii) spectral clustering. In the case of GMM, after computing the assignment probabilities, we assign each object to the cluster (distribution) with a maximal probability. We run each method and as well as correlation clustering (to obtain the ensemble solution) $100$ times and pick a solution with the smallest cost or negative log-likelihood.

\paragraph{\textbf{UCI datasets.}}
We perform our experiments on the following datasets selected randomly from the UCI data repository.\footnote{We observe similar results on several other datasets.}
\begin{enumerate}
\item \emph{Forest Type}:  contains multi-temporal  sensing information of $326$ samples from a forested area in Japan each described with $27$ features. The dataset consists of $5$ clusters.
\item \emph{Hayes-Roth}: contains $160$ samples on human subjects study each described with $5$ attributes.
\item \emph{Lung Cancer}: each instance contains $56$ attributes and is categorized as cancer or non-cancer.
\item \emph{Mammographic Mass}: consists of the BI-RADS attributes of the mammographic masses for $961$ samples.
\item \emph{One-Hundred Plant}: contains leaf samples for $100$ plant species for each $16$ samples with $64$ features ($1,600$ samples in total with $100$ clusters).
\item \emph{Perfume}: contains $560$ instances (odors) of $20$ different perfumes measured by a handheld odor meter.
\item \emph{Semeion Handwritten Digit}: features of $1593$ handwritten digits from around $80$  persons where each digit stretched in a rectangular box $16x16$ in a gray scale of $256$ values.
\item  \emph{Statlog (Australian Credit Approval)}: includes credit card data (described with $14$ attributes) of $690$ users.
\item \emph{Urban Land Cover}: contains $168$ high resolution aerial images  of $9$ types each represented by $148$ features.
\item \emph{Vertebral Column}: contains information of $6$ biomechanical features of $310$ patients categorized according to their status.
\end{enumerate}

In these datasets, the objects and as well as the features extracted from different dendrograms  are represented by vectors.  Thus, to obtain the pairwise distances, we compute the  squared Euclidean distances between the  respective vectors. Some clustering algorithms such as spectral clustering require pairwise similarities as input, instead of a vector-based representation. Therefore, as proposed in~\cite{mlChehreghani16}, we convert the pairwise distances $\mathbf X$ (or $\mathbf D^D$, if obtained from a dendrogram) to a similarity matrix $\mathbf S$ via $\mathbf S_{ij} = \max(\mathbf X) - \mathbf X_{ij} + \min(\mathbf X)$, where the $\max(.)$ and $\min(.)$ operations return the maximal and minimal elements of the given matrix. Note that an alternative transformation is an exponential function in the form of $\mathbf S_{ij} = \exp(-\frac{\mathbf X_{ij}}{\sigma^2})$, which requires fixing the free parameter $\sigma$ in advance. However, in particular in unsupervised learning, this task is nontrivial and the appropriates values of $\sigma$ occur in a very narrow range~\cite{Luxburg:2007}.

\paragraph{\textbf{Evaluation.}}
The  ground truth solutions of these datasets are available. Therefore, we can quantitatively measure the performance of each method by comparing the estimated and the true cluster labels. For each estimated clustering solution, we compute three commonly used quality measures: i) adjusted Mutual Information~\cite{Vinh:2010}, that gives the mutual information between the two estimated and true solutions, ii) adjusted Rand score~\cite{hubert1985comparing}, that computes the similarity between them,  and iii) V-measure~\cite{RosenbergH07}, that gives the harmonic mean of homogeneity and completeness. We compute the adjusted variants of these criteria, i.e., they yield zero for random solutions.

\paragraph{\textbf{Results.}}
Tables \ref{Table:results1} and \ref{Table:results2} show the results on different UCI datasets.  Each block row represents a separate dataset (in order, \emph{Forest Type}, \emph{Hayes-Roth}, \emph{Lung Cancer}, \emph{Mammographic Mass} and \emph{One-Hundred Plant} in Table \ref{Table:results1} and \emph{Perfume}, \emph{Semeion Handwritten Digit}, \emph{Statlog}, \emph{Urban Land Cover} and \emph{Vertebral Column} in Table \ref{Table:results2}). For each dataset, we investigate the different feature extraction methods (\emph{base}, PCA, LSA and those obtained by different dendrograms) with three different clustering algorithms. The goal of studying the three clustering algorithms is to demonstrate that our feature extraction methods can be used with various forms of clustering algorithms and are not limited to a specific algorithm. In this way, we investigate one probabilistic clustreing model (GMM), one which uses vector-based representation ($K$-means) and another that is applied to pairwise relations (spectral clustering).  The three evaluation criteria that we use are the most common criteria for evaluating clustering methods. The results of the ensemble method are shown in blue. For each clustering algorithm and each evaluation measure, the best result is bolded among the different feature extraction methods.

\begin{table*}[th!]
\caption{Permanence of different representations and clustering methods on different UCI  datasets. The five block rows correspond to the first five datasets, respectively to \emph{Forest Type}, \emph{Hayes-Roth}, \emph{Lung Cancer}, \emph{Mammographic Mass} and \emph{One-Hundred Plant}. The results of the ensemble method are shown in blue. For each clustering algorithm and each evaluation measure, the best result is bolded among the different feature extraction methods.
}
\label{Table:results1}
\begin{center}
\begin{small}
\begin{sc}
\begin{tabular}{|l|ccc|ccc|ccc|}
\hline
 &\multicolumn{3}{c|}{GMM}&\multicolumn{3}{c|}{K-means}&\multicolumn{3}{c|}{Spectral Clustering}\\

 Method & M.I. & Rand & V.M.  & M.I. & Rand & V.M. & M.I. & Rand & V.M.  \\
\hline
\hline
base	 	& 0.3897 & 0.3306 & 0.3959 & 0.5197 & 0.4987 & 0.5279 & 0.4380 & 0.4303 & 0.4438 \\
pca	 	&  0.3755 & 0.3496 & 0.4139 & 0.4742 & 0.4181 & 0.3453 & 0.4331 & 0.4170 & 0.4274 \\
lsa		& 0.3716 & 0.3472 & 0.3460 & 0.4633 & 0.4842 & 0.37 81 &  0.4484 & 0.4208 & 0.4513  \\
\hline
single	& 0.3544 & 0.3466 & 0.3592 & 0.3681 & 0.3321 & 0.3813 & 0.3705 & 0.3318 & 0.3838 \\
complete	& 0.3517 & 0.2792 & 0.3592 & 0.3517 & 0.2792 & 0.3592 & 0.3521 & 0.2785 & 0.3595  \\
average	& \bf{0.5294} & \bf{0.5316} & \bf{0.5370} & \bf{0.5294} & \bf{0.5316} & \bf{0.5370} & \bf{0.5325} & \bf{0.5235} & \bf{0.5417}  \\
ward		& 0.4718 & 0.3498 & 0.4812 & 0.4718 & 0.3498 & 0.4812 & 0.4718 & 0.3498 & 0.4812  \\
\hline
{\color{blue} ensemble}	& {\color{blue}0.4771} & {\color{blue}0.3661} & {\color{blue}0.4855} & {\color{blue}0.4752} & {\color{blue}0.3641} & {\color{blue}0.4838} & {\color{blue}0.4752} & {\color{blue}0.3641} & {\color{blue}0.4838}\\
\hline
\hline
base	 	& 0.1198 & 0.1175 & 0.1336 & 0.0138 & 0.0146 & 0.0005 & 0.0138 & 0.0146 & 0.0005  \\
pca	 	    & 0.1113 & 0.1096 & 0.1159 & 0.0376 & 0.0244 & 0.0107 & 0.0260 & 0.0205 & 0.0187 \\
lsa		    & 0.1375 & 0.1440 & 0.1425 & 0.0756 & 0.0517 & 0.0346 & 0.0878 & 0.1060 & 0.1250 \\
\hline
single	 &  \bf{0.2379} & 0.1624 & \bf{0.2589} & 0.2562 & 0.2035 & \bf{0.3561} & 0.2273 & \bf{0.1685} & \bf{0.2909} \\
complete	& 0.0446 & 0.0383 & 0.0588 & 0.0446 & 0.0383 & 0.0588 & 0.0446 & 0.0383 & 0.0588  \\
average	 & 0.1945 & \bf{0.1787} & 0.2118 & \bf{0.2610} & \bf{0.2403} & 0.2787 & \bf{0.2419} & 0.1614 & 0.2752 \\
ward		& 0.0249 & 0.0496 & 0.0412 & 0.0249 & 0.0496 & 0.0412 & 0.0249 & 0.0496 & 0.0412 \\
\hline
{\color{blue} ensemble}	& {\color{blue}0.1426} & {\color{blue}0.1193} & {\color{blue}0.1560} & {\color{blue}0.1249} & {\color{blue}0.1046} & {\color{blue}0. 1112} & {\color{blue}0.1042} & {\color{blue}0.1096} & {\color{blue}0.1152}\\
\hline
\hline
base	 	& 0.1684 & 0.1698 & 0.2030 & 0.1997 & 0.2294 & 0.2356 & 0.1197 & 0.1294 & 0.1356\\
pca	 	& 0.1170	& 0.1170 & 0.1743 & 0.1962 & 0.2362 & 0.2430 & 0.0609 & 0.0678 & 0.0890\\
lsa		& 0.1702 & 0.2162 & 0.2730 & 0.1962 & 0.2362 & 0.2430 & 0.0728 & 0.0419 & 0.0606\\
\hline
single	& 0.1677 & 0.2316 & 0.1892 & 0.1525 & 0.2636 & 0.2425 & 0.1016 & 0.1316 & 0.1282\\
complete	& 0.1537 & 0.2809 & 0.1810 & 0.1537 & 0.2809 & 0.1810 & 0.1537 & 0.2809 & 0.1810\\
average	& 0.1475 & 0.2253 & 0.1795 & 0.2070 & 0.3533 & 0.2303 & 0.1239 & 0.1327 & 0.0742\\
ward		& 0.1766 & 0.3388 & 0.2140 & 0.1766 & 0.3388 & 0.2140 & \bf{0.1766} & \bf{0.3388} & \bf{0.2140}\\
\hline
{\color{blue} ensemble}	& {\color{blue}\bf{0.2659}} & {\color{blue}\bf{0.4345}} & {\color{blue}\bf{0.2957}} & {\color{blue}\bf{0.2659}} & {\color{blue}\bf{0.4345}} & {\color{blue}\bf{0.2957}} & {\color{blue}\bf{0.1766}} & {\color{blue}\bf{0.3388}} & {\color{blue}\bf{0.2140}}\\
\hline
\hline
base	 	& 0.0036 & 0.0037 & 0.0059 & 0.0944 & 0.1133 & 0.0959 & 0.0944 & 0.1133 & 0.0959 \\
pca	    	& 0.0679 & 0.0454 & 0.0406 & 0.0944 & 0.1133 & 0.0959 & 0.0944 & 0.1133 & 0.0959   \\
lsa		    & 0.0550 & 0.0431 & 0.0603 & 0.0944 & 0.1133 & 0.0959 & 0.0944 & 0.1133 & 0.0959 \\
\hline
single	 & 0.0407 & 0.0915 & 0.0639 & \bf{0.1523} & \bf{0.2078} & \bf{0.1542} & \bf{0.1523} & \bf{0.2078} & \bf{0.1542}  \\
complete	& 0.0152 & 0.0113 & 0.0166 & 0.0152 & 0.0113 & 0.0166 & 0.0598 & 0.0191 & 0.0743  \\
average	 &  \bf{0.0834} & \bf{0.0721} & \bf{0.0895} & 0.0834 & 0.0721 & 0.0895 & 0.0834 & 0.0721 & 0.0895 \\
ward		& \bf{0.0834} & \bf{0.0721} & \bf{0.0895} & 0.0834 & 0.0721 & 0.0895 & 0.0834 & 0.0721 & 0.0895 \\
\hline
{\color{blue} ensemble}	& {\color{blue}\bf{0.0834}} & {\color{blue}\bf{0.0721}} & {\color{blue}\bf{0.0895}} & {\color{blue}0.0834} & {\color{blue}0.0721} & {\color{blue}0.0895} & {\color{blue}0.0834} & {\color{blue}0.0721} & {\color{blue}0.0895}\\
\hline
\hline
base		& 0.4834 & 0.1956 & 0.6867 & 0.6765 & 0.4844 & 0.8138 & 0.4386 & 0.2427 & 0.6547\\
pca		& 0.4510 & 0.2070 & 0.6791 & 0.6571 & 0.4580 & 0.8024 & 0.4704 & 0.2507 & 0.6881\\
lsa		& 0.4745 & 0.2942 & 0.7121 & 0.6593 & 0.4794 & 0.8034 & 0.4225 & 0.2185 & 0.6455\\
\hline
single	& 0.4841 & 0.2426 & 0.6915 & 0.4809 & 0.2354 & 0.6884 & 0.4922 & 0.2625 & 0.6982\\
complete	& 0.6381 & 0.4459 & 0.7893 & 0.6377 & 0.4427 & 0.7893 & 0.6377 & 0.4456 & 0.7891\\
average	& 0.6975 & 0.5336 & \bf{0.8258} & 0.6885 & 0.5176 & 0.8211 & 0.6788 & 0.5051 & 0.8159\\
ward		& 0.6914 & 0.5249 & 0.8207 & 0.6852 & 0.5158 & 0.8174 & \bf{0.6876} & 0.5151 & \bf{0.8184}\\
\hline					
{\color{blue}ensemble}	& {\color{blue}\bf{0.6990}} & {\color{blue}\bf{0.5408}} & {\color{blue}0.8251} & {\color{blue}\bf{0.6925}} & {\color{blue}\bf{0.5362}} & {\color{blue}\bf{0.8218}} & {\color{blue}0.6836} & {\color{blue}\bf{0.5197}} & {\color{blue}0.8164}\\
\hline
\end{tabular}
\end{sc}
\end{small}
\end{center}
\end{table*}

\begin{table*}[th!]
\caption{Permanence of different representations and clustering methods on different UCI  datasets. The five block rows correspond to the second five datasets, respectively to \emph{Perfume}, \emph{Semeion Handwritten Digit}, \emph{Statlog}, \emph{Urban Land Cover} and \emph{Vertebral Column}. The results of the ensemble method are shown in blue. For each clustering algorithm and each evaluation measure, the best result is bolded among the different feature extraction methods.
}
\label{Table:results2}
\begin{center}
\begin{small}
\begin{sc}
\begin{tabular}{|l|ccc|ccc|ccc|}
\hline
 &\multicolumn{3}{c|}{GMM}&\multicolumn{3}{c|}{K-means}&\multicolumn{3}{c|}{Spectral Clustering}\\

 Method & M.I. & Rand & V.M.  & M.I. & Rand & V.M. & M.I. & Rand & V.M.  \\
\hline
\hline
base		& 0.8350 & 0.6783 & 0.8944 & 0.8555 & 0.7243 & 0.8974 & 0.2353 & 0.3981 & 0.4070\\
pca		& 0.8916 & 0.7051 & 0.9159 & 0.8174 & 0.7430 & 0.8731 & 0.7933 & 0.6890 & 0.8942\\
lsa		& 0.7853 & 0.5912 & 0.8485 & 0.7982 & 0.6237 & 0.8625 & 0.8038 & 0.6049 & 0.8521\\
\hline
single	& 0.8975 & 0.7924 & 0.9178 & 0.8967 & 0.7939 & 0.9245 & 0.8943 & 0.7960 & 0.9213\\
complete	& 0.8941 & 0.7842 & 0.9169 & 0.8752 & 0.7474 & 0.9025 & 0.8632 & 0.7197 & 0.8981\\
average	& 0.9054 & 0.8193 & 0.9229 & 0.9116 & 0.8288 & 0.9298 & 0.9041 & 0.8088 & 0.9263\\
ward		& \bf{0.9390} & \bf{0.8831} & \bf{0.9516} & \bf{0.9348} & \bf{0.8729} & \bf{0.9491} & \bf{0.9348} & \bf{0.8729} & \bf{0.9491}\\
\hline
{\color{blue}ensemble}	& {\color{blue}0.9183} & {\color{blue}0.8393} & {\color{blue}0.9357} & {\color{blue}0.9133} & {\color{blue}0.8411} & {\color{blue}0.9379} & {\color{blue}0.9087} & {\color{blue}0.8244} & {\color{blue}0.9342}\\
\hline
\hline
base	 & 0.5253 & 0.4064 & 0.5312 & 0.5313 & 0.4037 & 0.5382 & 0.4884 & 0.3596 & 0.4970  \\
pca	 	&  0.5095 & 0.3685 & 0.5095 & 0.5291 & 0.4130 & 0.5179 & 0.4909 & 0.2928 & 0.5434 \\
lsa		&  0.5130 & 0.3619 & 0.5406 & 0.5217 & 0.4097 & 0.5226 & 0.4982 & 0.2641 & 0.4849 \\
\hline
single	&  0.4961 & 0.3806 & 0.5132 & 0.4943 & 0.3214 & 0.5258 & 0.5065 & 0.2740 & 0.5612 \\
complete	& 0.3911 & 0.2508 & 0.4010 & 0.4110 & 0.2780 & 0.4212 & 0.4206 & 0.2769 & 0.4349  \\
average	& \bf{0.5879} & \bf{0.4648} & 0.5740 & \bf{0.6004} & \bf{0.4712} & \bf{0.6132} & 0.5524 & 0.3682 & \bf{0.5896}  \\
ward		&  0.5362 & 0.3842 & 0.5495 & 0.5353 & 0.3770 & 0.5502 & 0.5587 & 0.3915 & 0.5736 \\
\hline
{\color{blue} ensemble}	& {\color{blue}0.5661} & {\color{blue}0.4214} & {\color{blue}\bf{0.5858}} & {\color{blue}0.5588} & {\color{blue}0.4211} & {\color{blue}0.5705} & {\color{blue}\bf{0.5759}} & {\color{blue}\bf{0.4181}} & {\color{blue}0.5863}\\
\hline
\hline
base		& 0.0074 & 0.0038 & 0.0162 & 0.0038 & 0.0022 & 0.0099 & 0.0232 & 0.0116 & 0.0425\\
pca		& 0.0074 & 0.0038 & 0.0162 & 0.0038 & 0.0022 & 0.0099 & 0.0525 & 0.0278 & 0.0261\\
lsa		& 0.0074 & 0.0038 & 0.0162 & 0.0074 & 0.0038 & 0.0162 & 0.0305 & 0.0374 & 0.0316\\
\hline
single	& 0.0580 & 0.0859 & 0.0593 & 0.0580 & 0.0859 & 0.0593 & 0.0219 & 0.0203 & 0.0357\\
complete	& 0.0399 & 0.0510 & 0.0411 & 0.0298 & 0.0445 & 0.0309 & 0.0570 & 0.0715 & 0.0709\\
average	& \bf{0.0864} & 0.1271 & \bf{0.0898} & 0.0367 & 0.0484 & 0.0476 & \bf{0.0719} & \bf{0.0972} & \bf{0.0830}\\
ward		& 0.0848 & 0.1251 & 0.0881 & \bf{0.0848} & \bf{0.1251} & \bf{0.0881} & 0.0074 & 0.0038 & 0.0162\\		
\hline					
{\color{blue}ensemble}	& {\color{blue}\bf{0.0864}} & {\color{blue}\bf{0.1272}} & {\color{blue}0.0896} & {\color{blue}\bf{0.0848}} & {\color{blue}\bf{0.1251}} & {\color{blue}\bf{0.0881}} & {\color{blue}0.0291} & {\color{blue}0.0259} & {\color{blue}0.0458}\\
\hline
\hline
base		& 0.1465 & 0.0844 & 0.1747 & 0.0909 & 0.0339 & 0.1277 & 0.1392 & \bf{0.0963} & 0.1645\\
pca		& 0.1465 & 0.0844 & 0.1747 & 0.0909 & 0.0339 & 0.1277 & 0.0705 & 0.0817 & 0.0763\\
lsa		& 0.1465 & 0.0844 & 0.1747 & 0.0909 & 0.0339 & 0.1277 & 0.1208 & 0.0953 & 0.1281\\
\hline
single	& 0.0973 & 0.0409 & 0.1236 & 0.0939 & 0.0458 & 0.1258 & 0.0898 & 0.0364 & 0.1272\\
complete	& \bf{0.1688} & \bf{0.0902} & \bf{0.1910} & \bf{0.1640} & \bf{0.0798} & \bf{0.1858} & \bf{0.1563} & 0.0689 & \bf{0.1769}\\
average	& 0.1489 & 0.0732 & 0.1708 & 0.1436 & 0.0659 & 0.1650 & 0.1493 & 0.0721 & 0.1711\\
ward		& 0.1515 & 0.0796 & 0.1746 & 0.1406 & 0.0594 & 0.1632 & 0.1406 & 0.0594 & 0.1632\\
\hline
{\color{blue}ensemble}	& {\color{blue}0.1534} & {\color{blue}0.0857} & {\color{blue}0.1771} & {\color{blue}0.1491} & {\color{blue}0.0756} & {\color{blue}0.1717} & {\color{blue}0.1501} & {\color{blue}0.0735} & {\color{blue}0.1724}\\
\hline
\hline
base	     	& 0.1159 & 0.0825 & 0.1257 & 0.2072 & 0.1051 & 0.1953 & \bf{0.1722} & 0.1042 & \bf{0.1779} \\
pca		        & 0.1398 & 0.06472 & 0.1534 & 0.1948 & 0.1601 & 0.1692 & 0.1209 & 0.1075 & 0.1383\\
lsa		        & 0.1308 & 0.1179 & 0.1445 & 0.1609 & 0.1388 & 0.1846 & 0.1630 & \bf{0.1252} & 0.1715\\
\hline
single	        & 0.1528 & 0.1002 & 0.1643 & 0.1906 & 0.2687 & 0.2001 & 0.1092 & 0.1188 & 0.1211 \\
complete	& 0.1696 & 0.1053 & 0.1773 & 0.1696 & 0.1053 & 0.1773 & 0.0705 & 0.0645 & 0.0941 \\
average  	& \bf{0.3080} & \bf{0.3278} & \bf{0.3247} & \bf{0.3080} & \bf{0.3278} & \bf{0.3247} & 0.1242 & 0.0560 & 0.1444 \\
ward		    & 0.1443 & 0.2216 & 0.1512 & 0.1443 & 0.2216 & 0.1512 & 0.1191 & 0.0583 & 0.1399 \\
\hline
{\color{blue}ensemble}	& {\color{blue}0.2475} & {\color{blue}0.2846} & {\color{blue}0.2613} & {\color{blue}0.2322} & {\color{blue}0.2776} & {\color{blue}0.2452} & {\color{blue}0.1165} & {\color{blue}0.0953} & {\color{blue}0.1376}\\
\hline
\end{tabular}
\end{sc}
\end{small}
\end{center}
\end{table*}


The \emph{base} method indicates performing the GMM, $K$-means or spectral clustering  on  the original vectors without inferring any new features. We also investigate Principal Component Analysis (PCA) and Latent Semantic Analysis (LSA) as two other baselines.
As discussed  in Theorem \ref{theorem:HierarchyEmbedding}, the matrix of pairwise dendrogram-based distances satisfy the \emph{ultrametric} conditions. Unltrametric is stronger than metric, i.e., any ultrametrci is a metric too. The only difference is the last condition in the proof of Theorem \ref{theorem:HierarchyEmbedding}. For an ultrametric, we require  $\forall i,j,k: \mathbf D^D_{ij}  \le  \max(\mathbf D^D_{ik},\mathbf D^D_{kj})$. It is obvious that this condition satisfies the triangle (metric) condition too, i.e.,   $\forall i,j,k: \mathbf D^D_{ij}  \le  \mathbf D^D_{ik} + \mathbf D^D_{kj}$. Hence, $\mathbf D^D$ induces a metric.
On the other hand, the different embedding methods usually rely on satisfying the metric conditions. Therefore, in principle any embedding and dimension reduction method can be applied to the dendrogram-based pairwise distances, the same way that it can be applied to the base pairwise distances too. Thus, further investigation of the results of different embedding methods is orthogonal to our contribution and we postpone it to future work.

Different dendrogram-based feature extraction methods are specified by the name of the criterion used to build the deprogram. The ensemble method refers to the aggregation of the different solutions and then preforming correlation clustering.
According to the equivalence of \emph{single} linkage method, Minimax distances and the tree preserving embedding method in \cite{ShiehHA11}, this method can be seen as another baseline which also constitutes a special instantiation of  the dendrogram-based feature extraction methodology.
Note that the superior performance of Minimax distances (\emph{single} linkage features) over methods such as metric learning or link-based methods has been demonstrated in previous works~\cite{KimC07icml,KimC13AAAI,ChehreghaniSDM16,ChehreghaniAAAI2017} (see for example Figure 1 in~\cite{KimC13AAAI}).
\footnote{Moreover, methods such as metric learning often require fixing free parameter(s) which is non-trivial in unsupervised settings such as clustering.}

We interpret the results of Tables \ref{Table:results1} and \ref{Table:results2} as follows. For each dataset (block row) and each clustering algorithm, we investigate whether ‘some’ of the dendrogram-based features (i.e., \emph{single}, \emph{complete}, \emph{average} or \emph{Ward}) perform better (according to the three evaluation criteria) than the baseline methods (\emph{base}, PCA and LSA). If so, then we conclude  our framework provides a rich and diverse family of non-parametric feature extraction methods wherein some instances yield more suitable features for the data at hand. Thus, a user has more freedom and options to choose the correct features.   However, the user might not have sufficient information to choose the correct features (dendrograms), thus, we propose to use the ensemble variant, in the context of averaging (aggregating) multiple learners.

According to the results reported in Tables \ref{Table:results1} and \ref{Table:results2}, we observe: i) extracting features from dendrograms yields better representations  that improve the evaluation scores of the final clusters. The dendrogram might be built in different ways which correspond to computing different types of features. In particular, we observe the features extracted via \emph{average} linkage and \emph{Ward} linkage often lead to very good results. \emph{Single} linkage (Minimax) features are more suitable for low-dimensional data wherein connectivity paths still exists. However, in higher dimensions, the other methods  might perform better due to robustness and flexibility. ii) The ensemble method works well in particular compared to the baselines and most of the dendrogram-based approaches. Note that the ensemble method is more than just averaging the results. It can be interpreted as obtaining a good (strong) learner from a set of weaker learners. Thereby, in several cases,  the ensemble method performs even better than all the other alternatives.

\paragraph{\textbf{Aggregation of representations.}}
As a side study, we investigate the sequential aggregation of different dendrogram-based features in representation space, i.e., we consider the combination of every two such  feature extractors. For this purpose, we first compute a dendrogram and extract the respective features. Then, we use these features to compute a second dendrogram from which we obtain a new set of  features. Finally, we apply a clustering method (GMM, $K$-means and spectral clustering) and evaluate the results  w.r.t. Mutual Information, Rand score and V-measure.

\begin{table}[th!]
\caption{Aggregation of two representations on the \emph{Perfume} dataset. The first and the second dendrograms are indicated by the rows and the columns, respectively. GMM is used to perform the clustering on the final features. The best combination is using first \emph{Ward} and then any of the four options.}
\label{Table:deep_GMM}
\vskip 0.15in
\begin{center}
\begin{small}
\begin{sc}
\begin{tabular}{|l|cccc|}
\hline

  & s & c & a  & w   \\
\hline
&\multicolumn{4}{c|}{M.I.} \\
\hline
s		& 0.9509 & 0.9120 & 0.8998 & 0.9182 \\
c	     & 0.8738 & 0.8787 & 0.9116 & 0.9246 \\
a		& 0.9305 & 0.9197 & 0.9305 & 0.9125 \\
w		& \bf{0.9612} & \bf{0.9612} & \bf{0.9612} & \bf{0.9612} \\
\hline
&\multicolumn{4}{c|}{Rand} \\
\hline
s		& 0.9071 & 0.8289 & 0.8114 & 0.8385 \\
c	& 0.7517 & 0.7622 & 0.8195 & 0.8443 \\
a		& 0.8678 & 0.8480 & 0.8678 & 0.8399 \\
w	& \bf{0.9360} & \bf{0.9360} & \bf{0.9360} & \bf{0.9360} \\
\hline
&\multicolumn{4}{c|}{V.M.} \\
\hline
s		& 0.9595 & 0.9255 & 0.9161 & 0.9325 \\
c	& 0.8991 & 0.9020 & 0.9302 & 0.9411 \\
a		& 0.9441 & 0.9350 & 0.9441 & 0.9263 \\
w		& \bf{0.9667} & \bf{0.9667} & \bf{0.9667} & \bf{0.9667} \\
\hline
\end{tabular}
\end{sc}
\end{small}
\end{center}
\end{table}

\begin{table}[th!]
\caption{Aggregation of two representations on the \emph{Perfume} dataset, where $K$-means is used for the clustering of the final features. W-S (\emph{Ward} and then \emph{single}) is the best combination.}
\label{Table:deep_Kmeans}
\vskip 0.15in
\begin{center}
\begin{small}
\begin{sc}
\begin{tabular}{|l|cccc|}
\hline
  & s & c & a  & w   \\
\hline
&\multicolumn{4}{c|}{M.I.} \\
\hline
s		& 0.9164 & 0.8945 & 0.9192 & 0.9104 \\
c	& 0.8666 & 0.8653 & 0.8842 & 0.9057 \\
a		& 0.9091 & 0.9091 & 0.9091 & 0.8911 \\
w	& \bf{0.9481} & 0.9383 & 0.9383 & 0.9379 \\
\hline
&\multicolumn{4}{c|}{Rand} \\
\hline
s		& 0.8341 & 0.7942 & 0.8416 & 0.8175 \\
c	& 0.7333 & 0.7332 & 0.7685 & 0.8061 \\
a		& 0.8225 & 0.8225 & 0.8225 & 0.7946 \\
w	& \bf{0.8963} & 0.8824 & 0.8824 & 0.8805 \\
\hline
&\multicolumn{4}{c|}{V.M.} \\
\hline
s		& 0.9313 & 0.9236 & 0.9342 & 0.9300 \\
c	& 0.9019 & 0.8999 & 0.9145 & 0.9267 \\
a		& 0.9295 & 0.9295 & 0.9295 & 0.9116 \\
w		& \bf{0.9603} & 0.9516 & 0.9516 & 0.9515 \\
\hline
\end{tabular}
\end{sc}
\end{small}
\end{center}
\end{table}

\begin{table}[th!]
\caption{Aggregation of two representations on the \emph{Perfume} dataset, where spectral clustering is applied to the final features to cluster them. W-S (\emph{Ward}-\emph{single}) is the best combination.}
\label{Table:deep_SP}
\vskip 0.15in
\begin{center}
\begin{small}
\begin{sc}
\begin{tabular}{|l|cccc|}
\hline
  & s & c & a  & w   \\
\hline
&\multicolumn{4}{c|}{M.I.} \\
\hline
s		& 0.9147 & 0.8768 & 0.8832 & 0.9104 \\
c	& 0.8373 & 0.8339 & 0.8538 & 0.8717 \\
a		& 0.8578 & 0.8460 & 0.8458 & 0.8756 \\
w		& \bf{0.9217} & 0.9161 & 0.9124 & 0.9139 \\
\hline
&\multicolumn{4}{c|}{Rand} \\
\hline
s		& 0.8366 & 0.7643 & 0.7788 & 0.8175 \\
c	& 0.6875 & 0.6736 & 0.7165 & 0.7412 \\
a		& 0.7070 & 0.6958 & 0.6901 & 0.7448 \\
w	& \bf{0.8368} & 0.8327 & 0.8339 & 0.8350 \\
\hline
&\multicolumn{4}{c|}{V.M.} \\
\hline
s 		& 0.9366 & 0.9093 & 0.9133 & 0.9300 \\
c	& 0.8836 & 0.8819 & 0.8972 & 0.9075 \\
a		& 0.9005 & 0.8922 & 0.8905 & 0.9107 \\
w	& \bf{0.9426} & 0.9364 & 0.9367 & 0.9375 \\
\hline
\end{tabular}
\end{sc}
\end{small}
\end{center}
\end{table}

\begin{table*}[th!]
\caption{Comparison of \emph{Ward}(W) and \emph{Ward-single}(W-S) features on the \emph{perfume} dataset. Performing \emph{single} linkage on the \emph{Ward} features improves the final clustering.
}
\label{Table:Perfume_comparison}
\begin{center}
\begin{small}
\begin{sc}
\begin{tabular}{|l|ccc|ccc|ccc|}
\hline
 &\multicolumn{3}{c|}{GMM}&\multicolumn{3}{c|}{K-means}&\multicolumn{3}{c|}{Spectral Clustering}\\

 Method & M.I. & Rand & V.M.  & M.I. & Rand & V.M. & M.I. & Rand & V.M.  \\
\hline\hline
W	 	& 0.9390 & 0.8831 & 0.9516 & 0.9348 & 0.8729 & 0.9491 & \bf{0.9348} & \bf{0.8729} & \bf{0.9491}\\
W-S     	& \bf{0.9612}	& \bf{0.9360} & \bf{0.9667} & \bf{0.9481} & \bf{0.8963} & \bf{0.9603} & 0.9217 & 0.8368 & 0.9426\\
\hline
\end{tabular}
\end{sc}
\end{small}
\end{center}
\end{table*}

We observe for most of the datasets, aggregation of different features either  improves the results or preserves the accuracy of the results as  same as the first representation. However, aggregation of the clustering solutions usually yields more significant changes (improvements) compared to  the aggregating the representations. One of the  significant changes happens on the  \emph{Perfume} dataset. See the results in Tables \ref{Table:deep_GMM}, \ref{Table:deep_Kmeans}  and \ref{Table:deep_SP}, where respectively GMM, $K$-means and spectral clustering have been applied to the final features to produce the clusters. The first and the second dendrograms are indicated by the rows and the columns, respectively (where S refers to \emph{single}, C to \emph{complete}, A to \emph{average}, and W to \emph{Ward}, the different ways of obtaining the features).
These results should be compared with the block row in Table \ref{Table:results2} that corresponds to the \emph{Perfume} dataset (the first block row). We observe that over this dataset, feature aggregation often improves the results for different clustering methods. However, as mentioned before, such an aggregation is usually less significant (on other datasets).

We observe that on this dataset, the W-S combination (extracting the features first via \emph{Ward} and then via \emph{single} linkages)  consistently yields the best results, among all different combinations.
In Table \ref{Table:Perfume_comparison}, we compare these results with the best feature extractor for the \emph{perfume} dataset, which is based on the \emph{Ward} linkage. \emph{Single} linkage even though does not yield very good results itself, but improves the \emph{Ward} features the most. According to Table \ref{Table:Perfume_comparison}, except spectral clustering, using the \emph{single} linkage features helps the clustering algorithm to produce better results. However, the best result is obtained with GMM for which combining \emph{Ward} with any option is helpful.

\paragraph{\textbf{Model selection.}}
Our framework provides several options for choosing the dendrogram and the level function, and at the same time a principled way to aggregate and choose the best options (either in solution space or in representation space).

Availability of such  alternatives endows a rich family of unsupervised models for representation learning and feature extraction. We note that this availability is different than optimizing the free parameters of a kernel.
\begin{enumerate}
  \item In our framework, the number of the choices is very limited, whereas for a kernel function the free parameter(s) can usually take a wide (continuous) range of different values. Moreover,  the optimal values of the kernel parameters usually occur   inside very narrow ranges that makes it difficult to find them  via search or cross-validation, even using the labeled data \cite{Nadler07,Luxburg:2007}.
  \item In our framework, every choice has an explicit interpretation that makes model selection more straightforward. For example, single linkage is more suitable for elongated structures and patterns, whereas average linkage suits better for high-dimensional data. On the other hand,   the proposed level function in Eq. \ref{eq:level_hierarchy_dist} is better adapted to the density-diverse structures.
  \item Finally, as we demonstrated on all the datasets, our framework also provides a consistent way for  computing an ensemble of the different choices and options. According to the experimental results, the ensemble solution performs very well compared to the individual choices. Computing such an ensemble solution is nontrivial for many kernels.
\end{enumerate}

 Here, as a side study, we compare the two choices for level function on the ensemble solution, i.e., the option defined in Eq. \ref{eq:singleMinimax} and the one defined in Eq. \ref{eq:level_hierarchy_dist}.  As explained before,  Eq. \ref{eq:level_hierarchy_dist} suggests  a context-sensitive level function that takes into account the data diversity.
 According to the results in Tables \ref{Table:results1} and  \ref{Table:results2}, with the level function in Eq. \ref{eq:level_hierarchy_dist}, the ensemble solution of GMM on different UCI datasets yields the following MI scores: 0.4771, 0.1426, 0.2659, 0.0834, 0.6990, 0.9183, 0.5661, 0.0864, 0.1534, 0.2475. However, with the level function in  Eq. \ref{eq:singleMinimax}, the ensemble solution of GMM on different UCI datasets gives the following MI scores:  0.4148, 0.1301, 0.2738, 0.0834, 0.6364, 0.9075, 0.5747, 0.0864, 0.1451, 0.2262. We observe that on two datasets (\emph{Mammographic Mass} and \emph{Statlog}) the two variants yield the same results. Among the remaining eight datasets, on six of them the level function in Eq. \ref{eq:level_hierarchy_dist} performs better, whereas on only two datasets (\emph{Lung Cancer} and \emph{Semeion Handwritten Digit}) the level function in Eq. \ref{eq:singleMinimax} yields higher scores. It is notable that however the results from both of the choices are acceptable.

\paragraph{\textbf{Efficiency of correlation clustering optimization.}}
In our framework, we employ an efficient optimization of correlation clustering to compute the ensemble solution.  We have studied its effectiveness in terms of the quality of the ensemble solution. Here, we investigate the efficiency of its optimization procedure  in terms of runtimes. In particular, we compare our local search optimization with the Linear Programming (LP) method  \cite{DemaineEFI06} and the Semidefinite Programming relaxation (SDP)  \cite{CharikarGW03,MathieuS10}. Table \ref{Table:runtime_comparison} shows the different runtime results. We observe that the local search method performs significantly faster compared to the alternatives. It is notable that the SDP method encounters memory issues for the datasets of larger than $200$ objects.  We stop it when its runtime exceeds $10$ hours.

\begin{table*}[th!]
\caption{Comparison of the runtimes of different methods for optimization the correlation clustering objective used to obtain the ensemble solutions.}
\label{Table:runtime_comparison}
\begin{center}
\begin{tabular}{|l|ccc|}
\hline
 dataset & local search & LP  & SDP   \\
\hline
\emph{Forest Type}	 	& 2.8 sec & 149.3 sec &  more than 10 hours\\
\emph{Hayes-Roth}   	& 2.1  sec & 93.0 sec & 8.45 hours\\
\emph{Lung Cancer}   & 1.1 sec & 26.7 sec  & 2.16 hours \\
\emph{Mammographic Mass}    	& 16.5 sec & 729.5 sec & more than 10 hours \\
\emph{One-Hundred Plant}      	& 27.2 sec &  1015.9 sec & more than 10 hours \\
\emph{Perfume}        	& 4.7 sec & 312.9 sec &  more than 10 hours  \\
\emph{Semeion Handwritten Digit}          	&  4.9 sec & 335.7 sec  & more than 10 hours  \\
\emph{Statlog}            	& 6.2 sec & 369.8 sec  & more than 10 hours \\
\emph{Urban Land Cover}    	& 2.2 sec & 95.2 sec & 9.51 hours\\
\emph{Vertebral Column}      	& 2.6 sec  & 127.8 sec & more than 10 hours  \\
\hline
\end{tabular}
\end{center}
\end{table*}

\paragraph{\textbf{Experiments on  scientific datasets.}}

\begin{table*}[th!]
\caption{Permanence of different representations and clustering methods on two  scientific datasets. The first block row corresponds to the \emph{computer science} dataset and the second block row corresponds to \emph{electrical engineering} dataset. The results of the ensemble method are shown in blue. For each clustering algorithm and each evaluation measure, the best result is bolded among the different feature extraction methods.
}
\label{Table:results3}
\begin{center}
\begin{small}
\begin{sc}
\begin{tabular}{|l|ccc|ccc|ccc|}
\hline
 &\multicolumn{3}{c|}{GMM}&\multicolumn{3}{c|}{K-means}&\multicolumn{3}{c|}{Spectral Clustering}\\

 Method & M.I. & Rand & V.M.  & M.I. & Rand & V.M. & M.I. & Rand & V.M.  \\
\hline
\hline
base	          & 0.3291 & 0.3136 & 0.4065 & 0.3530 & 0.3264 & 0.3168 & 0.3705 & 0.3530 & 0.3822 \\
pca		          & 0.3205 & 0.3021 & 0.3437 & 0.3474 & 0.3628 & 0.3340 & 0.3606 & 0.3502 & 0.3797 \\
lsa		          & 0.3066 & 0.3139 & 0.3839 & 0.3522 & 0.3328 & 0.3874 & 0.3430 & 0.3588 & 0.3678  \\
\hline
single      	& 0.3628 & 0.3750 & 0.4196 & 0.3965 & 0.3921 & 0.4391 & 0.3729 & 0.3780 & 0.4125  \\
complete	& 0.5043 & 0.3951 & 0.5784 & 0.4966 & 0.3902 & 0.5557 & 0.4831 & 0.3877 & 0.5484  \\
average 	& \bf{0.5067} & \bf{0.4193} & \bf{0.5728} & \bf{0.7007} & \bf{0.7423} & \bf{0.7196} & \bf{0.6241} & \bf{0.6825} & \bf{0.6654} \\
ward		   & 0.4638 & 0.3404 & 0.5381 & 0.4972 & 0.3753 & 0.5688 & 0.4535 & 0.3260 & 0.5212 \\
\hline
{\color{blue}ensemble}	& {\color{blue}0.4726} & {\color{blue}0.3687} & {\color{blue}0.5447} & {\color{blue}0.4956} & {\color{blue}0.4032} & {\color{blue}0.5669} & {\color{blue}0.4949} & {\color{blue}0.3728} & {\color{blue}0.5456}\\
\hline
\hline
base	 &  0.2718 & 0.1528 & 0.3204 & 0.2652 & 0.1502 & 0.3077 & 0.2883 & 0.2040 & 0.3184 \\
pca	 	&   0.2744 & 0.1475 & 0.3254 & 0.2683 & 0.1434 & 0.3114 & 0.2818 & 0.2159 & 0.3401 \\
lsa		&   0.2803 & 0.1614 & \bf{0.3390} & 0.2413 & 0.1332 & 0.2932 & 0.2743 & 0.2115 & 0.3217\\
\hline
single	        & \bf{0.2915} & 0.1683 &0.3372& \bf{0.2949} & 0.1655 & \bf{0.3367} & 0.3132 & 0.2006 & 0.3485   \\
complete	& 0.2907 & 0.1819 & 0.3362 & 0.2576 & 0.1588 & 0.3022 & 0.2913 & 0.2265 & 0.3625 \\
average    	& 0.2354 & 0.1419 & 0.2804 & 0.2848 & \bf{0.1725} & 0.3303 & 0.2987 & 0.1940 & 0.3321  \\
ward		    & 0.2810 & \bf{0.1878} & 0.3354 & 0.2451 & 0.1518 & 0.2827 & 0.2852 & 0.1960 & 0.3229 \\
\hline
{\color{blue} ensemble}	& {\color{blue}0.2779} & {\color{blue}0.1653} & {\color{blue}0.3277} & {\color{blue}0.2234} & {\color{blue}0.1495} & {\color{blue}0.2670} & {\color{blue}\bf{0.3348}} & {\color{blue}\bf{0.2273}} & {\color{blue}\bf{0.3666}}\\
\hline
\end{tabular}
\end{sc}
\end{small}
\end{center}
\end{table*}

At the end, we investigate the proposed methods on two real-world datasets collected within a scientific data analytics project. The goal is to extract clusters of different subjects and topics. The extracted clusters help to analyze i) how an automated approach can distinguish the scientific outcomes in different subjects and accordingly categorize the respective authors, ii) how  separable or related the different subjects and topics are.
The first dataset contains $10,000$ published scientific articles in $10$ different topics of computer science including algorithms, database, machine learning, networks, hardware, software engineering, formal methods, security, logic and information systems.  The second dataset contains $10,000$ published scientific articles in  different topics of electrical engineering. Each ground truth cluster consists of $1,000$ articles.  For each dataset, we obtain the TF-IDF vectors of the articles where we remove the step words.  The number of features is $5,823$ and $5,495$ respectively  for computer science and for electrical engineering datasets.
We compute the base pairwise distances based on the squared Euclidean distances between the TF-IDF vectors. We then use them to compute the dendrograms.

Table \ref{Table:results3} shows the permanence of different representations and clustering methods on these  datasets, where the first block row corresponds to the \emph{computer science} dataset and the second block row corresponds to the \emph{electrical engineering} dataset. The results of the ensemble method are shown in blue. For each clustering algorithm and each evaluation measure, the best result is bolded among the different feature extraction methods. We observe consistent results to the UCI datasets. i) Using different dendrogram-based features often improves the results for different clustering methods w.r.t. the evaluation criteria. ii) The ensemble solution yields either  the best results or yields  very close results to the best choice, i.e., it can effectively address the model selection problem.

\section{Conclusion}
We extended the previous  Minimax and tree preserving  representation learning methods that  correspond to building a \emph{single} linkage dendrogram, and proposed a generic framework to compute representations from different dendrograms, beyond \emph{single} linkage. Then, we studied the embedding to extract vector-based features for such distances.
This property extends the applicability to a wide range of machine learning algorithms. Then, we considered the aggregation of different dendrogram-based features in solution space and representation space.
First, based on the consistency of the cluster labels of different objects, we build a graph with positive and negative edge weights and then apply correlation clustering to obtain the final clusters. In the second approach, in the spirit of deep learning models, we apply different dendrogram-based features sequentially, such that the input of the next layer is the output of the current one, and then we apply the particular (clustering) algorithm to the final features.
Our experiments on several datasets revealed the effectiveness of the proposed framework.

\section*{Acknowledgement}
This work was partially supported by the Wallenberg AI, Autonomous Systems and Software Program (WASP) funded by the Knut and Alice Wallenberg Foundation.
Parts of this work have been done at Xerox Research.

\bibliographystyle{plain}
\bibliography{references}

\end{document}